\documentclass[runningheads]{llncs}
\usepackage{graphicx} 
\usepackage{hyperref}  
\usepackage{relsize}  
\usepackage{cprotect}  
\usepackage[dvipsnames]{xcolor} 
\usepackage{comment}  
\usepackage{amssymb} 
\usepackage{ulem} 
\usepackage[font=small]{caption}

\usepackage{setspace} 

\usepackage{caption}
\usepackage{mathtools}
\DeclareMathOperator*{\argmax}{arg\,max}

\begin{document}
\begin{sloppypar}
\title{A Hierarchical Model for Data-to-Text Generation}

%
%
\author{\mbox{Cl\'ement Rebuffel\inst{1,2} \and
Laure Soulier\inst{1} \and
Geoffrey Scoutheeten\inst{2} \and
Patrick Gallinari\inst{1,3}}}
%
%
\institute{LIP6, Sorbonne Universit\'e, France 
\and
BNP Paribas, France
\and
Criteo AI Lab, Paris \\
\email{firstname.lastname@\{lip6.fr,bnpparibas.com\}}}
\maketitle              
\begin{abstract}
Transcribing structured data into natural language descriptions has emerged as a challenging task, referred to as ``data-to-text". These structures generally regroup multiple elements, as well as their attributes. Most attempts rely on translation encoder-decoder methods which linearize elements into a sequence. This however loses most of the structure contained in the data. In this work, we propose to overpass this limitation with a hierarchical model that  encodes the data-structure at the element-level and the structure level. Evaluations on  RotoWire show the effectiveness of our model w.r.t. qualitative and quantitative metrics. 
\end{abstract}

\section{Introduction}
\label{section:introduction}

\begin{figure}
    \makebox[\textwidth][c]{\includegraphics[width=1.1\textwidth]{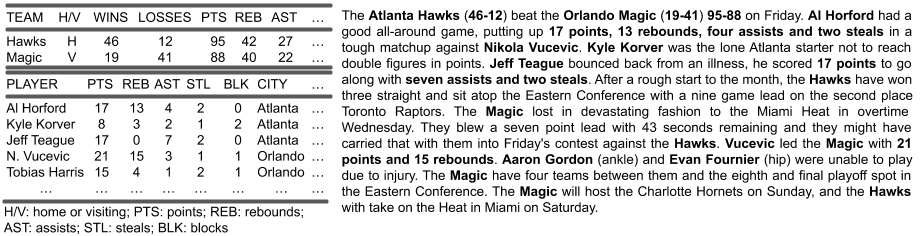}}%
    \vspace{-0.2cm}
    \caption{Example of structured data from the RotoWire dataset. Rows are entities (either a team or a player) and each cell a record, its key being the column label and its value the cell content. Factual mentions from the table are boldfaced in the description.}
    \vspace{-0.6cm}
    \label{fig:rotowire-ex}
\end{figure}

Knowledge and/or data is often modeled in a structure, such as indexes, tables, key-value pairs, or triplets. 
These data, by their nature (e.g., raw data or long time-series data), are not easily usable by humans; outlining their crucial need to be synthesized.
Recently, numerous works have focused on leveraging structured data in various applications, such as question answering \cite{pasupat2015,sun2016} or table retrieval \cite{deng2019,DasSarma2012}. One emerging research field consists in transcribing data-structures into natural language in order to ease their understandablity and their usablity. This field is referred  to as ``data-to-text" \cite{gatt2018} and has its place in several application domains (such as journalism \cite{oremus2014} or medical diagnosis \cite{pauws2018}) or wide-audience applications (such as financial \cite{Plachouras2016} and weather reports \cite{reiter2005}, or sport broadcasting \cite{chen2008,wiseman2017}).
As an example, Figure \ref{fig:rotowire-ex} shows a data-structure containing statistics on NBA basketball games, paired with its corresponding journalistic description. 

Designing data-to-text models gives rise to two main challenges: 1) understanding structured data and 2) generating associated descriptions. 
Recent data-to-text models \cite{liu2018,puduppully2018,puduppully2019,wiseman2017} mostly rely on an encoder-decoder architecture \cite{bahdanau2014} in which the data-structure is first encoded sequentially into a fixed-size vectorial representation by an encoder. Then, a decoder generates words conditioned on this representation. With the introduction of the attention mechanism \cite{luong2015} on one hand, which computes a context focused on important elements from the input at each decoding step and, on the other hand, the copy mechanism \cite{gulcehre2016,see2017} to deal with unknown or rare words, these systems produce fluent and domain comprehensive texts.
For instance, Roberti et al. \cite{roberti2019} train a character-wise encoder-decoder to generate descriptions of restaurants based on their attributes, while Puduppully et al. \cite{puduppully2018} design a more complex two-step decoder: they first generate a plan of elements to be mentioned, and then condition text generation on this plan.
Although previous work yield overall good results, we identify two important caveats, that hinder precision (\textit{i.e.} factual mentions) in the descriptions: 
\vspace{-0.2cm}
\begin{enumerate}
    \item \textit{Linearization of the data-structure.} In practice, most works focus on introducing innovating decoding modules, and still represent data as a unique sequence of elements to be encoded. For example, the table from Figure \ref{fig:rotowire-ex} would be linearized to [(Hawks, H/V, H), ..., (Magic, H/V, V), ...], effectively leading to losing distinction between rows, and therefore entities. To the best of our knowledge, only Liu et al. \cite{liu2019,liu2018} propose encoders constrained by the structure but these approaches are designed for single-entity structures. 
    \item \textit{Arbitrary ordering of unordered collections in recurrent networks (RNN).} Most data-to-text systems use RNNs as encoders (such as GRUs or LSTMs), these architectures have however some limitations. Indeed, they require in practice their input to be fed sequentially. This way of encoding unordered sequences (\textit{i.e.} collections of entities) implicitly assumes an arbitrary order within the collection which, as demonstrated by Vinyals et al. \cite{vinyals2016},
    significantly impacts the learning performance.
\end{enumerate}

To address these shortcomings, we propose a new structured-data encoder assuming that structures should be hierarchically captured. Our contribution focuses on the encoding of the data-structure, thus the decoder is chosen to be a classical module as used in \cite{puduppully2018,wiseman2017}. 
Our contribution is threefold:
\vspace{-0.2cm}
\begin{itemize}
    \item We model the general structure of the data using a two-level architecture, first encoding all entities on the basis of their elements, then encoding the data structure on the basis of its entities;
    \item We introduce the Transformer encoder \cite{vaswani2017} in data-to-text models to ensure robust encoding of each element/entities in comparison to all others, no matter their initial positioning;
    \item We integrate a hierarchical attention mechanism to compute the hierarchical context fed into the decoder.
\end{itemize}

We report experiments on the  RotoWire benchmark  \cite{wiseman2017} which contains around $5K$ statistical tables of NBA basketball games paired with human-written descriptions.
Our model is compared to several state-of-the-art models. Results show that the proposed architecture outperforms previous models on BLEU score and is generally better 
on qualitative metrics.

In the following, we first present a state-of-the art of data-to-text  literature (Section 2), and then describe our proposed hierarchical data encoder (Section~3). The evaluation protocol is presented in Section 4, followed by the  results (Section~5). Section 6 concludes the paper and presents  perspectives.
\vspace{-0.2cm}

\section{Related Work}
\label{section:related-work}

Until recently, efforts to bring out semantics from structured-data relied heavily on expert knowledge \cite{deng2013,reiter2005}. For example,  
in order to better transcribe numerical time series of weather data to a textual forecast, Reiter et al. \cite{reiter2005} devise complex template schemes in collaboration with weather experts to build a consistent set of data-to-word rules.

Modern approaches to the wide range of tasks based on structured-data (\textit{e.g.} table retrieval \cite{deng2019,zhang2019-table}, table classification \cite{ghasemi-gol2018}, question answering \cite{haug2018}) now propose to leverage progress in deep learning to represent these data into a semantic vector space (also called embedding space). 
In parallel, an emerging task, called ``data-to-text", aims at describing structured data into a natural language description. This task stems from the neural machine translation (NMT) domain, and early work \cite{agarwal2017,lebret2016,wiseman2017} represent the data records as a single sequence of facts to be entirely translated into natural language. Wiseman et al. \cite{wiseman2017} show the limits of traditional NMT systems on larger structured-data, where NMT systems fail to accurately extract salient elements.

To improve these models, a number of work \cite{li2018,puduppully2018,wiseman2018} proposed innovating decoding modules based on planning and templates, to ensure factual and coherent mentions of records in generated descriptions. For example, Puduppully et al. \cite{puduppully2018} propose a two-step decoder which first targets specific records and then use them as a plan for the actual text generation. Similarly, Li et al. \cite{li2018} proposed a delayed copy mechanism where their decoder also acts in two steps: 1) using a classical LSTM decoder to generate delexicalized text  and 2) using a pointer network \cite{vinyals2015} to replace placeholders by records from the input data.

Closer to our work, very recent work \cite{liu2018,liu2019,puduppully2019} have  proposed to take into account the data structure. More particularly, Puduppully et al. \cite{puduppully2019} follow entity-centric theories \cite{grosz1995,mann1988} and propose a model based on dynamic entity representation at decoding time. It consists in conditioning the decoder on entity representations that are updated during inference at each decoding step. On the other hand, Liu et al. \cite{liu2018,liu2019} rather focus on introducing structure into the encoder. For instance, they propose a dual encoder \cite{liu2019} which encodes separately the sequence of element names and the sequence of element values. These approaches are however designed for single-entity data structures and do not account for delimitation between entities.

Our contribution differs from previous work in several aspects. 
First, instead of flatly concatenating elements from the data-structure and encoding them as a sequence \cite{liu2018,puduppully2018,wiseman2017}, we constrain the encoding  to the underlying structure of the input data, so that the delimitation between entities remains clear throughout the process.
Second,  unlike all works in the domain, we exploit the Transformer architecture \cite{vaswani2017} and leverage its particularity to directly compare elements with each others in order to avoid arbitrary assumptions on their ordering.   
Finally, in contrast to \cite{clark2018,puduppully2019} that use a complex updating mechanism to obtain a dynamic representation of the input data and its entities, we argue that explicit hierarchical encoding naturally guides the decoding process via hierarchical attention.

\vspace{-0.1cm}
\section{Hierarchical Encoder Model for Data-to-Text}
\label{section:model}
\vspace{-0.1cm}
In this section we introduce our proposed hierarchical model taking into account the data structure.
We outline that the decoding component aiming to generate  descriptions is considered as a black-box module  so that our contribution is focused on  the encoding module.
We first describe the model overview, before detailing the hierarchical encoder and the associated hierarchical attention. 

\vspace{-0.2cm}
\subsection{Notation and General Overview}
\label{subsection:overview}
\vspace{-0.1cm}
Let's consider the following notations:

$\bullet$ An \textit{entity} $e_i$ is a set of $J_i$ unordered records $\{r_{i,1}, ..., r_{i,j}, ..., r_{i,J_i}\}$; where  record $r_{i,j}$ is defined as a pair of \textit{key} $k_{i,j}$ and \textit{value} $v_{i,j}$.  We outline that $J_i$ might differ between entities.

$\bullet$  A \textit{data-structure} $s$ is an unordered set of $I$ entities $e_i$. We thus denote $s \coloneqq \{e_1, ..., e_i, ..., e_I\}$.

$\bullet$ For each data-structure, a textual \textit{description} $y$ is associated. We  refer to the first $t$ words of a description $y$ as $y_{1:t}$. Thus, the full sequence of words can be noted as $y = y_{1:T}$.

$\bullet$ The \textit{dataset} $\mathcal{D}$ is a collection of $N$ aligned (data-structure, description) pairs $(s,y)$. \newline 
For instance, Figure \ref{fig:rotowire-ex} illustrates a data-structure associated with a description. The data-structure includes a set of entities (\textit{Hawks, Magic, Al Horford, Jeff Teague, ...}). The entity  Jeff Teague is modeled as a set of records \{(PTS, 17), (REB, 0), (AST, 7) ...\} in which, e.g., the record (PTS, 17) is characterized by a \textit{key} (PTS) and a \textit{value} (17). \newline

For each data-structure $s$ in $\mathcal{D}$, the objective function aims to generate a description  $\hat{y}$ as close as possible to the ground truth $y$.
This objective function optimizes the following log-likelihood over the whole dataset $\mathcal{D}$:

\begin{equation} \label{eq:log-likelihood}
    \argmax_\theta \mathcal{L}(\theta) = \argmax_\theta \sum_{(s,y) \in \mathcal{D}}{\log P(\hat{y}=y\ |\ s; \theta)}
\end{equation} 
\noindent where $\theta$ stands for the model parameters and $P(\hat{y}=y\ |\ s; \theta)$ the probability of the model to generate the adequate description $y$ for table $s$.

During inference, we generate the sequence $\hat{y}^*$ with the maximum a posteriori probability conditioned on  table $s$. Using the chain rule, we get:
\vspace{-0.2cm}
\begin{equation} \label{eq:inference}
    \hat{y}_{1:T}^* = \argmax_{\hat{y}_{1:T}} \prod_{t=1}^T P(\hat{y}_t | \hat{y}_{1:t-1}; s; \theta)
\end{equation}

This  equation  is intractable in practice, we approximate a solution using beam search, as in \cite{liu2018,liu2019,puduppully2018,puduppully2019,wiseman2017}.

Our model follows the  encoder-decoder architecture \cite{bahdanau2014}. Because our contribution focuses  on the encoding process, we chose the decoding module used in \cite{puduppully2018,wiseman2017}: a two-layers LSTM network with a copy mechanism. In order to supervise this mechanism, we assume that each record value that also appears in the target  is copied from the  data-structure and we train the model to switch between freely generating words from the vocabulary and copying words from the input.
We now describe the hierarchical encoder and the hierarchical attention.

\subsection{Hierarchical Encoding Model}
\label{subsection:hierarchy}

\begin{figure}[t]
    \centering
    \makebox[\textwidth][c]{\includegraphics[scale=0.6]{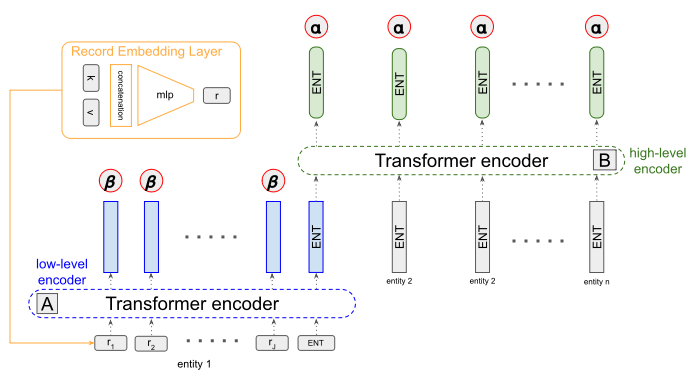}}%
    \vspace{-0.4cm}
    \cprotect\caption{\begin{small}Diagram of the proposed hierarchical encoder. Once the records are embedded, the low-level encoder works on each entity independently (A); then the high-level encoder encodes the collection of entities (B). In circles, we represent the hierarchical attention scores: the $\alpha$ scores at the entity level and the $\beta$ scores at the record level.\end{small}}
    \vspace{-0.4cm}
    \label{fig:model}
\end{figure}

As outlined in Section \ref{section:related-work}, most previous work \cite{li2018,puduppully2018,puduppully2019,wiseman2017,wiseman2018} make use of flat encoders that do not exploit the data structure.
To keep the semantics of each element from the data-structure, we propose a  hierarchical encoder which relies on two modules. The first one (module A in Figure \ref{fig:model}) is called \textit{low-level encoder} and encodes entities on the basis of their records; the second one (module B), called \textit{high-level encoder}, encodes the data-structure on the basis of its underlying entities.
In the low-level encoder, the traditional embedding layer is replaced by a record embedding layer as in  \cite{liu2018,puduppully2018,wiseman2017}. 
We present in what follows the record embedding layer  and introduce our two hierarchical modules.

\subsubsection{Record Embedding Layer.}
\label{subsubsection:record-encoder}
The first layer of the network consists in learning two embedding matrices to embed the record keys and values. Keys $k_{i,j}$ are embedded to $\mathbf{k}_{i,j} \in \mathbb{R}^{d}$ and values $v_{i,j}$ to $\mathbf{v}_{i,j} \in \mathbb{R}^{d}$, with $d$ the size of the embedding.  As in previous work \cite{liu2018,puduppully2018,wiseman2017}, each record embedding $\mathbf{r}_{i,j}$ is computed by a linear projection on the concatenation $[\mathbf{k}_{i,j}$;\ $\mathbf{v}_{i,j}]$ followed by a non linearity:
\begin{equation} \label{eq:record-embedding}
    \mathbf{r}_{i,j} = \text{ReLU}(\mathbf{W}_{r}[\mathbf{k}_{i,j};\ \mathbf{v}_{i,j}] + \mathbf{b}_{r})
    \vspace{-0.2cm}
\end{equation}
\noindent where $\mathbf{W}_r \in \mathbb{R}^{2d \times d}$ and $\mathbf{b}_r \in \mathbb{R}^{d}$ are learnt parameters.

The low-level encoder aims at encoding a collection of records belonging to the same entity while the high-level encoder encodes the whole set of entities.
Both the low-level and high-level encoders consider their input elements as unordered. We use the Transformer architecture from \cite{vaswani2017}.
For each encoder, we have the following peculiarities:
\vspace{-0.2cm}
\begin{itemize}
    \item the \textbf{Low-level encoder} encodes each entity $e_i$ on the basis of its  record embeddings $\mathbf{r}_{i,j}$.
    Each record embedding $\mathbf{r}_{i,j}$ is compared to other record embeddings to learn its final hidden representation $\mathbf{h}_{i,j}$.
    Furthermore, we add a special record \verb|[ENT]| for each entity, illustrated in Figure \ref{fig:model} as the last record. Since entities might have a variable number of records, this token allows to aggregate final hidden record representations $\{\mathbf{h}_{i,j}\}_{j=1}^{J_i}$  in a fixed-sized  representation vector $\mathbf{h}_{i}$. 
    \item the \textbf{High-level encoder} encodes the data-structure on the basis of its entity representation $\mathbf{h}_{i}$. Similarly to the \textbf{Low-level encoder}, the final hidden state $\mathbf{e_i}$ of an entity is computed by comparing entity representation $\mathbf{h}_{i}$ with each others. The data-structure representation $\mathbf{z}$ is computed as the mean of these entity representations, and is used for the decoder initialization.
\end{itemize}

\subsection{Hierarchical attention} 
\label{subsubsection:attn}

To fully leverage the hierarchical structure of our encoder, we propose two variants of hierarchical attention mechanism to compute the context fed to the decoder module.

$\bullet$ \textit{Traditional Hierarchical Attention.} As in \cite{puduppully2019}, we hypothesize that a dynamic context should be computed in two steps: first attending to entities, then to records corresponding to these entities. To implement this hierarchical attention, at each decoding step $t$, the model learns a first set of attention scores $\alpha_{i,t}$ over entities $e_i$ and a second set of attention scores $\beta_{i,j,t}$ over records $r_{i,j}$ belonging to  entity $e_i$. The $\alpha_{i,t}$ scores are normalized to form a distribution over all entities $e_i$, and $\beta_{i,j,t}$ scores are normalized to form a distribution over records $r_{i,j}$ of  entity $e_i$. Each entity is then represented as a weighted sum of its record embeddings, and the entire data structure is represented as a weighted sum of the entity representations. 
The dynamic context is computed as:
\vspace{-0.2cm}
\begin{align}
 &\mathbf{c_t} = \sum_{i=1}^{I} (\alpha_{i,t} \big(  \sum_{j} \beta_{i,j,t} \mathbf{r}_{i,j} \big)) \label{eq:context} \\
 where ~~ \alpha_{i,t} &\propto exp(\mathbf{d}_t\mathbf{W}_\alpha\mathbf{e}_i)  ~~and~~
\beta_{i,j,t} \propto exp(\mathbf{d}_t\mathbf{W}_\beta\mathbf{h}_{i,j})\label{eq:hierarchical-kv}
\end{align}

\noindent where $\mathbf{d_t}$ is the decoder hidden state at time step $t$, $\mathbf{W}_{\alpha} \in \mathbb{R}^{d\times d}$ and $\mathbf{W}_{\beta} \in \mathbb{R}^{d\times d}$ are learnt parameters, $ \sum_i\alpha_{i,t} = 1$, and for all $i \in \{1,...,I\}$  $\sum_{j}\beta_{i,j,t} = 1$.

$\bullet$ \textit{Key-guided Hierarchical Attention.}
This variant follows the intuition that once an entity is chosen for mention (thanks to $\alpha_{i,t}$), only the type of records is important to determine the content of the description.
For example, when deciding to mention a player, all experts automatically report his score without consideration of its specific value. To test this intuition, 
we model the attention scores by computing the $\beta_{i,j,t}$ scores from equation (\ref{eq:hierarchical-kv}) solely on the embedding of the  \textit{key} rather than on the full record representation $\mathbf{h}_{i,j}$: 
\vspace{-0.2cm}
\begin{equation} \label{eq:hierarchical-k}
    \hat{\beta}_{i,j,t} \propto exp(\mathbf{d}_t\mathbf{W}_{a_2}\mathbf{k}_{i,j})
    \vspace{-0.2cm}
\end{equation}

Please note that the different embeddings and the model parameters presented in the model components are learnt using Equation 1.

\section{Experimental setup}
\label{section:xp}

\subsection{The Rotowire dataset}

To evaluate the effectiveness of our model, and demonstrate its flexibility at handling heavy data-structure made of several types of entities, we used the RotoWire dataset \cite{wiseman2017}. It includes basketball games statistical tables paired with journalistic descriptions of the games, as can be seen in the example of  Figure~\ref{fig:rotowire-ex}. The descriptions are professionally written and average 337 words with a vocabulary size of $11.3$K. There are 39 different record keys, and the average number of records (resp. entities) in a single data-structure is 628 (resp. 28). Entities are of two types, either team or player, and player descriptions depend on their involvement in the game. We followed the data partitions introduced with the dataset and used a train/validation/test sets of respectively $3,398$/$727$/$728$ (data-structure, description) pairs.

\subsection{Evaluation metrics}
We evaluate our model through two types of metrics. The BLEU score \cite{papineni2002} aims at measuring to what extent the generated descriptions are literally closed to the ground truth. The second category designed by \cite{wiseman2017} is more qualitative.

\paragraph{BLEU Score.} The \textbf{BLEU score} \cite{papineni2002} is commonly used as an evaluation metric in text generation tasks. It estimates the correspondence between a machine output and that of a human by computing the number of co-occurrences for ngrams ($n \in {1, 2, 3, 4}$) between the generated candidate and the ground truth. 
We use the implementation code released by \cite{post2018}.

\paragraph{Information extraction-oriented metrics.} These metrics estimate the ability of our model to integrate elements from the table in its descriptions. Particularly, they compare the gold and generated descriptions and measure to what extent the extracted relations are aligned or differ. To do so, we follow the protocol presented in \cite{wiseman2017}. First, we apply an information extraction (IE) system trained on labeled relations from the gold descriptions of the RotoWire train dataset. Entity-value pairs are extracted from the descriptions. For example, in the sentence \textit{Isaiah Thomas led the team in scoring, totaling 23 points [...].},  an IE tool will extract the pair (Isaiah Thomas, 23, PTS). 
Second, we compute three metrics on the extracted information:\\
$\bullet$ \textbf{Relation Generation (RG)} estimates how well the system is able to generate text containing factual (i.e., correct) records. We measure the precision and absolute number (denoted respectively RG-P\% and RG-\#) of unique relations $r$ extracted from $\hat{y}_{1:T}$ that also appear in $s$. \\
$\bullet$ \textbf{Content Selection (CS)} measures how well the generated document matches the gold document in terms of mentioned records. We measure the precision and recall (denoted respectively CS-P\% and CS-R\%) of unique relations $r$ extracted from $\hat{y}_{1:T}$ that are also extracted from $y_{1:T}$.\\ 
$\bullet$ \textbf{Content Ordering (CO)} analyzes how well the system orders the records  discussed in the description. We measure the normalized Damerau-Levenshtein distance \cite{brill2000} between the sequences of records extracted from $\hat{y}_{1:T}$ that are also extracted from $y_{1:T}$. 

\vspace{-0.1cm}
CS primarily targets the ``what to say" aspect of evaluation, CO targets the ``how to say it" aspect, and RG targets both. 
Note that for CS, CO, RG-\% and BLEU metrics, higher is better; which is not true for RG-\#. The IE system used in the experiments is able to extract an average of $17$ factual records from gold descriptions. In order to mimic a human expert, a generative system should  approach this number and not overload generation with brute facts.

\subsection{Baselines}
We compare our hierarchical model against three systems. For each of them, we report the results of the best performing models presented in each paper.

$\bullet$ \textit{Wiseman}  \cite{wiseman2017} is a standard encoder-decoder system with copy mechanism.

$\bullet$ \textit{Li} \cite{li2018} is a standard encoder-decoder with a delayed copy mechanism: text is first generated with placeholders, which are replaced by salient records extracted from the table by a pointer network.

$\bullet$ \textit{Puduppully-plan} \cite{puduppully2018} acts in two steps: a first standard encoder-decoder generates a plan, \textit{i.e.} a list of salient records from the table; a second standard encoder-decoder generates text from this plan.

$\bullet$ \textit{Puduppully-updt}  \cite{puduppully2019}. It consists in a standard encoder-decoder, with an added module aimed at updating record representations during the generation process. At each decoding step, a gated recurrent network computes which records should be updated and what should be their new representation.

\vspace{-0.2cm}
\subsubsection{Model scenarios} 
We test the importance of the input structure by training different variants of the proposed architecture:

$\bullet$ \textit{Flat}, where we feed the input sequentially to the encoder, losing all notion of hierarchy. As a consequence, the model uses standard attention. This variant is closest to \textit{Wiseman}, with the exception that we use a Transformer to encode the input sequence instead of an RNN.

$\bullet$ \textit{Hierarchical-kv} is our full hierarchical model, with traditional hierarchical attention, \textit{i.e.} where attention over records is computed on the full record encoding, as in equation (\ref{eq:hierarchical-kv}). 

$\bullet$ \textit{Hierarchical-k} is our full hierarchical model, with key-guided hierarchical attention, \textit{i.e.} where attention over records is computed only on the record key representations, as in equation (\ref{eq:hierarchical-k}).

\subsection{Implementation details}
\vspace{-0.2cm}
The decoder is the one used in  \cite{puduppully2018,puduppully2019,wiseman2017} with the same hyper-parameters. For the encoder module, both the low-level and high-level encoders use a two-layers multi-head self-attention with two heads. To fit with the small number of record keys in our dataset (39), their embedding size is fixed to 20. The size of the record value embeddings and hidden layers of the Transformer encoders are both set to 300.
We use dropout at rate 0.5. The models are trained with a batch size of 64. We follow the training procedure in \cite{vaswani2017} and train the model for a fixed number of 25K updates, and average the weights of the last 5 checkpoints (at every 1K updates) to ensure more stability across runs. All models were trained with the Adam optimizer \cite{kingma2014}; the initial learning rate is 0.001, and is reduced by half every 10K steps. We used beam search with beam size of 5 during inference. All the models are implemented in OpenNMT-py \cite{opennmt}. All code is available at \href{https://github.com/KaijuML/data-to-text-hierarchical}{https://github.com/KaijuML/data-to-text-hierarchical}

\section{Results}
\vspace{-0.2cm}

\begin{table}[t]
\centering
\def\arraystretch{1}  
    \begin{tabular}{l|c|cc|ccc|c|c}
    \hline
    & BLEU & \multicolumn{2}{c|}{RG} & \multicolumn{3}{c|}{CS} & CO & Nb     \\
    &      & P\%        & \#         & P\%        & R\%  &F1       &      & Params\\
    \hline
    Gold descriptions               & 100  & 96.11      & 17.31      & 100        & 100   &100     & 100 &        \\
    Wiseman        & 14.5 & 75.62      & $\mathbf{16.83}$      & 32.80      & 39.93   &36.2   & 15.62 &  45M    \\
     Li        & 16.19 & 84.86      & $19.31$      & 30.81      & 38.79   &34.34   & 16.34 &  -    \\
    Pudupully-plan & 16.5 & 87.47      & 34.28      & 34.18      & 51.22  & 41     & 18.58 & 35M     \\
    Puduppully-updt  & 16.2 & $\mathbf{92.69}$      & 30.11      & 38.64      & 48.51  & 43.01    & $\mathbf{20.17}$ &  23M    \\
                       &      &            &            &            &            &       &      \\
    Flat            & $16.7_{.2}$ & $76.62_{1}$      & $18.54_{.6}$      & $31.67_{.7}$    & $42.9_{1}$ & $36.42_{.4}$     & $14.64_{.3}$  & 14M      \\
    Hierarchical-kv    & $17_{.3}$ & $89.04_{1}$      & $21.46_{.9}$      & $38.57_{1.2}$    & $51.50_{.9}$ &   $44.19_{.7}$  & $18.70_{.7}$  &  14M     \\
    Hierarchical-k     & $\mathbf{17.5}_{.3}$ & $89.46_{1.4}$      & $21.17_{1.4}$      & $\mathbf{39.47}_{1.4}$    & $\mathbf{51.64}_{1}$  &$\mathbf{44.7}_{.6}$    & $18.90_{.7}$  & 14M       \\  \hline
    \end{tabular}
\caption{Evaluation on the RotoWire testset using relation generation (RG) count (\#) and precision (P\%), content selection (CS) precision (P\%) and recall (R\%), content ordering (CO), and BLEU. -: number of parameters  unavailable.}
\vspace{-0.8cm}
\label{table:results}
\end{table}

Our results on the RotoWire testset are summarized in Table \ref{table:results}. For each proposed variant of our architecture, we report the mean score over ten runs, as well as the standard deviation in subscript. Results are compared to baselines \cite{puduppully2018,puduppully2019,wiseman2017} and variants of our models. We also report the result of the oracle (metrics on the gold descriptions). Please note that gold descriptions trivially obtain 100\% on all metrics expect RG, as they are all based on comparison with themselves. RG scores are different, as the IE system is imperfect and fails to extract accurate entities 4\% of the time. RG-\# is an absolute count.

\paragraph{Ablation studies}
To evaluate the impact of our model components, we first compare scenarios \textit{Flat}, \textit{Hierarchical-k}, and \textit{Hierarchical-kv}.  As shown in Table \ref{table:results}, we can see the lower results obtained by the \textit{Flat} scenario compared to the other scenarios (\textit{e.g.} BLEU $16.7$ vs. $17.5$ for resp. \textit{Flat} and \textit{Hierarchical-k}), suggesting the effectiveness of encoding the data-structure using a hierarchy. This is expected, as losing explicit delimitation between entities makes it harder  a) for the encoder to encode semantics of the objects contained in the table and b) for the attention mechanism to extract salient entities/records.

\begin{figure}[t]
    \makebox[\textwidth][c]{\includegraphics[width=1\textwidth]{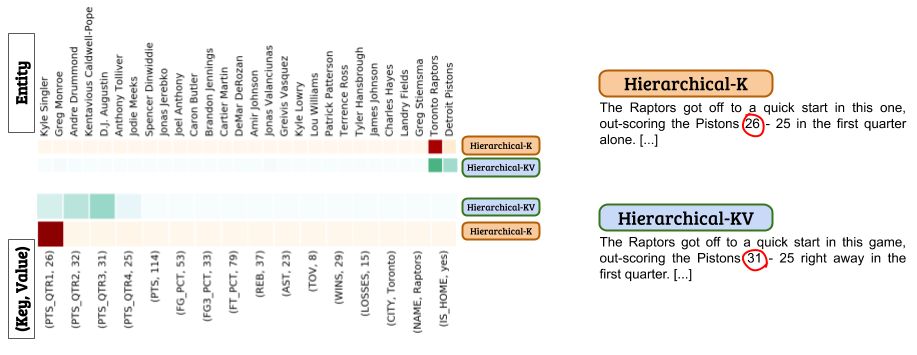}}%
    \vspace{-0.4cm}
    \caption{Right: Comparison of a generated sentence from \textit{Hierarchical-k} and \textit{Hierarchical-kv}. Left: Attention scores over entities (top) and  over records inside the  selected entity (bottom) for both variants, during the decoding of respectively 26 or 31 (circled in red).}
    \vspace{-0.2cm}
    \label{fig:attn}
\end{figure}

Second, the comparison between scenario \textit{Hierarchical-kv} and  \textit{Hierarchical-k} shows that omitting entirely the influence of the record values in the attention mechanism is more effective: this last variant performs slightly better in all metrics excepted CS-R\%, reinforcing our intuition that focusing on the structure modeling is an important part of data encoding as well as confirming the intuition explained in Section \ref{subsubsection:attn}: once an entity is selected, facts about this entity are relevant based on their key, not value which might add noise.
To illustrate this intuition, we depict in Figure \ref{fig:attn} attention scores (recall $\alpha_{i,t}$ and $\beta_{i,j,t}$ from equations (\ref{eq:hierarchical-kv}) and (\ref{eq:hierarchical-k})) for both variants \textit{Hierarchical-kv} and  \textit{Hierarchical-k}. We particularly focus on the timestamp where the models should mention the number of points scored during the first quarter of the game. Scores of \textit{Hierarchical-k} are sharp, with all of the weight on the correct record (PTS\_QTR1, 26) whereas scores of \textit{Hierarchical-kv} are more distributed over all PTS\_QTR records, ultimately failing to retrieve the correct one.

\vspace{-0.3cm}
\paragraph{Comparison w.r.t. baselines.}
From a general point of view, we can see from Table~\ref{table:results} that our scenarios obtain significantly higher results in terms of BLEU over all models; our best model \textit{Hierarchical-k} reaching $17.5$ vs. $16.5$ against the best baseline. This means that our models learns to generate fluent sequences of words, close to the gold descriptions, adequately picking up on domain lingo. Qualitative metrics are either better or on par with baselines.
We show in Figure \ref{fig:example-text} a text generated by our best model, which can be directly compared to the gold description in Figure \ref{fig:rotowire-ex}. Generation is fluent and contains domain-specific expressions. As reflected in Table \ref{table:results}, the number of correct mentions (in green) outweights the number of incorrect mentions (in red). Please note that, as in previous work \cite{li2018,puduppully2018,puduppully2019,wiseman2017}, generated texts still contain a number of incorrect facts, as well hallucinations (in blue): sentences that have no basis in the input data (e.g. \textit{``[...] he's now averaging 22 points [...]."}).  While not the direct focus of our work, this highlights that any operation meant to enrich the semantics of structured data can also enrich the data with incorrect facts.

\begin{figure}[t]
    {\scriptsize \begin{spacing}{0.5}
The \textbf{Atlanta Hawks} ( \textcolor{OliveGreen}{46 - 12} ) defeated the \textbf{Orlando Magic} ( \textcolor{OliveGreen}{19 - 41} ) \textcolor{OliveGreen}{95 - 88} on Monday at Philips Arena in Atlanta. The \textbf{Hawks} got out to a quick start in this one, out - scoring the \textbf{Magic} \textcolor{OliveGreen}{28 - 16 in the first quarter alone}. Along with the quick start, the \textbf{Hawks} were able to hold off the \textbf{Magic} late in the fourth quarter, out - scoring the \textbf{Magic} \textcolor{OliveGreen}{19 - 21}. The \textcolor{Red}{\textbf{Hawks}} were led by \textbf{Nikola Vucevic}, who went \textcolor{OliveGreen}{10 - for - 16 from the field} and \textcolor{OliveGreen}{0 - for - 0 from the three-point line} to score a team - high of \textcolor{OliveGreen}{21 points}, while also adding \textcolor{OliveGreen}{15 rebounds} in \textcolor{OliveGreen}{37 minutes}. \textcolor{BlueViolet}{It was his second double - double in a row, a stretch where he's averaging 22 points and 17 rebounds}. Notching a double - double of his own, \textbf{Al Horford} recorded \textcolor{OliveGreen}{17 points} ( \textcolor{OliveGreen}{7 - 9 FG , 0 - 0 3Pt , 3 - 4 FT} ), \textcolor{OliveGreen}{13 rebounds} and \textcolor{Red}{four steals}. \textcolor{BlueViolet}{He's now averaging 15 points and 6 rebounds on the year}. \textbf{Paul Millsap} had a strong showing , posting \textcolor{OliveGreen}{20 points} ( \textcolor{OliveGreen}{8 - 17 FG , 4 - 7 3Pt , 0 - 2 FT} ), \textcolor{OliveGreen}{four rebounds} and \textcolor{Red}{three blocked shots}. He's been a pleasant surprise for the \textbf{Magic} in the second half, as \textcolor{BlueViolet}{he's averaged 14 points and 5 rebounds over his last three games}. \textbf{DeMarre Carroll} was the other starter in double figures, finishing with \textcolor{OliveGreen}{15 points} ( \textcolor{OliveGreen}{6 - 12 FG , 3 - 6 3Pt} ), \textcolor{OliveGreen}{eight rebounds} and \textcolor{OliveGreen}{three steals}. \textcolor{BlueViolet}{He's had a nice stretch of three games , averaging 24 points, 3 rebounds and 2 assists over that span}. \textbf{Tobias Harris} was the only other \textbf{Magic} player to reach double figures, scoring \textcolor{OliveGreen}{15 points} ( \textcolor{OliveGreen}{5 - 9 FG , 2 - 4 3Pt , 3 - 4 FT} ). \textcolor{BlueViolet}{The \textbf{Magic} 's next game will be at home against the Miami Heat on Wednesday, while the \textbf{Magic} will travel to Charlotte to play the Hornets on Wednesday}. \end{spacing}}

    \caption{Text generated by our best model. Entites are boldfaced, factual mentions are in green, erroneous mentions in red and hallucinations are in blue.}
    \vspace{-0.5cm}
    \label{fig:example-text}
\end{figure}

Specifically, regarding all baselines, we can outline the following statements.
\indent $\bullet$ Our hierarchical models achieve significantly better scores on all metrics when compared to the flat architecture \textit{Wiseman}, reinforcing the crucial role of structure in data semantics and saliency. The analysis of RG metrics shows that  \textit{Wiseman} seems to be the more naturalistic in terms of number of  factual mentions (RG\#) since it is the closest scenario to the gold value (16.83 vs. 17.31 for resp. \textit{Wiseman} and \textit{Hierarchical-k}).  However, \textit{Wiseman} achieves only $75.62$\% of precision, effectively mentioning on average a total of $22.25$ records (wrong or accurate), where our model \textit{Hierarchical-k} scores a precision of $89.46$\%, leading to $23.66$ total mentions, just slightly above \textit{Wiseman}.

$\bullet$ The comparison between the \textit{Flat} scenario and  \textit{Wiseman} is particularly interesting.  Indeed, these two models share the same intuition to flatten the data-structure. The only difference stands on the encoder mechanism: bi-LSTM vs. Transformer, for \textit{Wiseman} and \textit{Flat} respectively. 
Results shows that our \textit{Flat} scenario  obtains a significant higher BLEU score (16.7 vs. 14.5) and generates fluent descriptions with accurate mentions (RG-P\%) that are also included in the gold descriptions (CS-R\%). This suggests that introducing the Transformer architecture is promising way to implicitly account for data structure.

\indent $\bullet$  Our hierarchical   models outperform the two-step decoders of \textit{Li} and \textit{Puduppully-plan} on both BLEU and all qualitative metrics, showing that capturing structure in the encoding process is more effective that predicting a structure in the decoder (i.e., planning or templating). While our models sensibly outperform in precision at factual mentions, the baseline \textit{Puduppully-plan} reaches $34.28$ mentions on average, showing that incorporating modules dedicated to entity extraction leads to over-focusing on entities; contrasting with our models that learn to generate more balanced descriptions.

\indent $\bullet$  The comparison with \textit{Puduppully-updt} shows that dynamically updating the encoding across the generation process can lead to  better Content Ordering (CO) and RG-P\%. However, this does not help with Content Selection (CS) since our best model \textit{Hierarchical-k} obtains slightly better scores. 
Indeed, \textit{Puduppully-updt} updates representations after each mention  allowing to keep track of the mention history. This  guides the ordering of mentions (CO metric), each step limiting more the number of candidate mentions (increasing RG-P\%). In contrast, our model encodes saliency among records/entities more effectively (CS metric). We note that while our model encodes the data-structure once and for all, \textit{Puduppully-updt} recomputes, via the updates, the encoding at each step and therefore significantly increases computation complexity. Combined with their RG-\# score of $30.11$, we argue that  our model is simpler, and obtains fluent description with accurate mentions in a more human-like fashion.

We would also like to draw attention to the number of parameters used by those architectures. We note that our scenarios relies on a lower number of parameters (14 millions) compared to all baselines (ranging from 23 to 45 millions). This outlines the effectiveness in the design of our model relying on a structure encoding, in contrast to other approach that try to learn the structure of data/descriptions from a linearized encoding.

\vspace{-0.2cm}
\section{Conclusion and future work}
\vspace{-0.2cm}
In this work we have proposed a hierarchical encoder for structured data, which 1)~leverages the structure to form efficient representation of its input; 2)~has strong synergy with the hierarchical attention of its associated decoder. This results in an effective and more light-weight model. Experimental evaluation on the RotoWire benchmark shows that our model outperforms competitive baselines in terms of BLEU score and is generally better on qualitative metrics. This way of representing structured databases may lead to automatic inference and enrichment, e.g., by comparing entities. This direction could be driven by very recent operation-guided networks \cite{trask2018,nie2018}. In addition, we note that our approach can still lead to erroneous facts or even hallucinations. An interesting perspective might be to further constrain the model on the data structure in order to prevent inaccurate of even contradictory descriptions.

\section{Acknowledgements}
\vspace{-0.2cm}
We would like to thank the H2020 project AI4EU (825619) which partially supports Laure Soulier and Patrick Gallinari.
%
%
%
\bibliographystyle{splncs04}
\bibliography{biblio}
\end{sloppypar}
\end{document}